\documentclass{article}

\PassOptionsToPackage{numbers, compress}{natbib}

    \usepackage[preprint]{tackling_climate_workshop_style}

\usepackage[utf8]{inputenc} 
\usepackage[T1]{fontenc}    
\usepackage{hyperref}       
\usepackage{url}            
\usepackage{booktabs}       
\usepackage{amsfonts}       
\usepackage{nicefrac}       
\usepackage{microtype}      
\usepackage{graphicx}       
\usepackage{float}          
\usepackage{subcaption}     

\title{Mapping of Land Use and Land Cover (LULC) using EuroSAT and Transfer Learning}

\author{%
  Suman Kunwar \\
  Faculty of Computer Science\\
  Selinus University of Sciences and Literature\\
  Ragusa, Italy \\
  \texttt{sumn2u@gmail.com} \\
   \And
  Jannatul Ferdush \\
 Deparment of Computer Science and Engineering\\
 Jashore University of Science and Technology \\
 Jashore, Bangladesh\\
\texttt{jannatulkuet@gmail.com} \\
}

\begin{document}

\maketitle

\begin{abstract}
As the global population continues to expand, the demand for natural resources increases. Unfortunately, human activities account for $23$\% of greenhouse gas emissions. On a positive note, remote sensing technologies have emerged as a valuable tool in managing our environment. These technologies allow us to monitor land use, plan urban areas, and drive advancements in areas such as agriculture, climate change mitigation, disaster recovery, and environmental monitoring. Recent advances in AI, computer vision, and earth observation data have enabled unprecedented accuracy in land use mapping. By using transfer learning and fine-tuning with RGB bands, we achieved an impressive $99.19$\% accuracy in land use analysis. Such findings can be used to inform conservation and urban planning policies.
\end{abstract}

\section{Introduction}
The world population has increased significantly over the last few centuries and is projected to continue growing \cite{mathieu_what_2023}. With growing demand, human beings are consuming more natural resources, including water, energy, minerals, and agricultural products. Activities such as agriculture, forestry, and urbanization contribute to $23$\% of global greenhouse gas emissions \cite{levin_7_2019}, primarily due to deforestation and land degradation. Monitoring land use changes is vital for better environmental management, urban planning, and nature protection \cite{huang_land_2021,banchero_recent_2020}.

With advancements in remote sensing technologies, satellite image data are now either freely available or can be commercially acquired, promoting innovation and entrepreneurship. Thanks to the use of low orbit and geostationary satellites \cite{zhou_patternnet:2018}, we can observe the Earth with unprecedented detail. Moreover, improvements in remote sensing technology have resulted in better spatial resolution \cite{usgs_role:2023}, enabling more precise ground surface analyses. Such data access has fueled advancements in agriculture, urban development, climate change mitigation, disaster recovery, and environmental monitoring \cite{hao_attention-based_2021, sarath_effect_2020, alex_review_2023}. Advances in computer vision, AI, and earth observation data facilitate large-scale land use mapping \cite{gladys_villegas_mapping_2021,sathyanarayanan_multiclass_2020}. The community has extensively embraced methods for classifying land use and land cover (LULC), including machine learning \cite{waghela_land_2022} and deep learning (DL) \cite{alem_deep_2020}. Recent studies suggest that DL techniques demonstrate remarkable performance in remote sensing (RS) image scene classification \cite{9887201}. The objective of image scene classification and retrieval is to automatically allocate class labels to every RS image scene stored in an archive. This differs from semantic segmentation tasks used for LULC mapping and classification.

One application of this technology is scene classification \cite{cheng_remote_2020}, which involves labeling an image based on specific semantic categories. This approach has numerous practical uses, including LULC analysis as well as land resource management \cite{karra_global_2021, ojha_land_2019}.

\begin{figure}[H]
  \centering
  \includegraphics[width=0.97\textwidth, height=8cm]{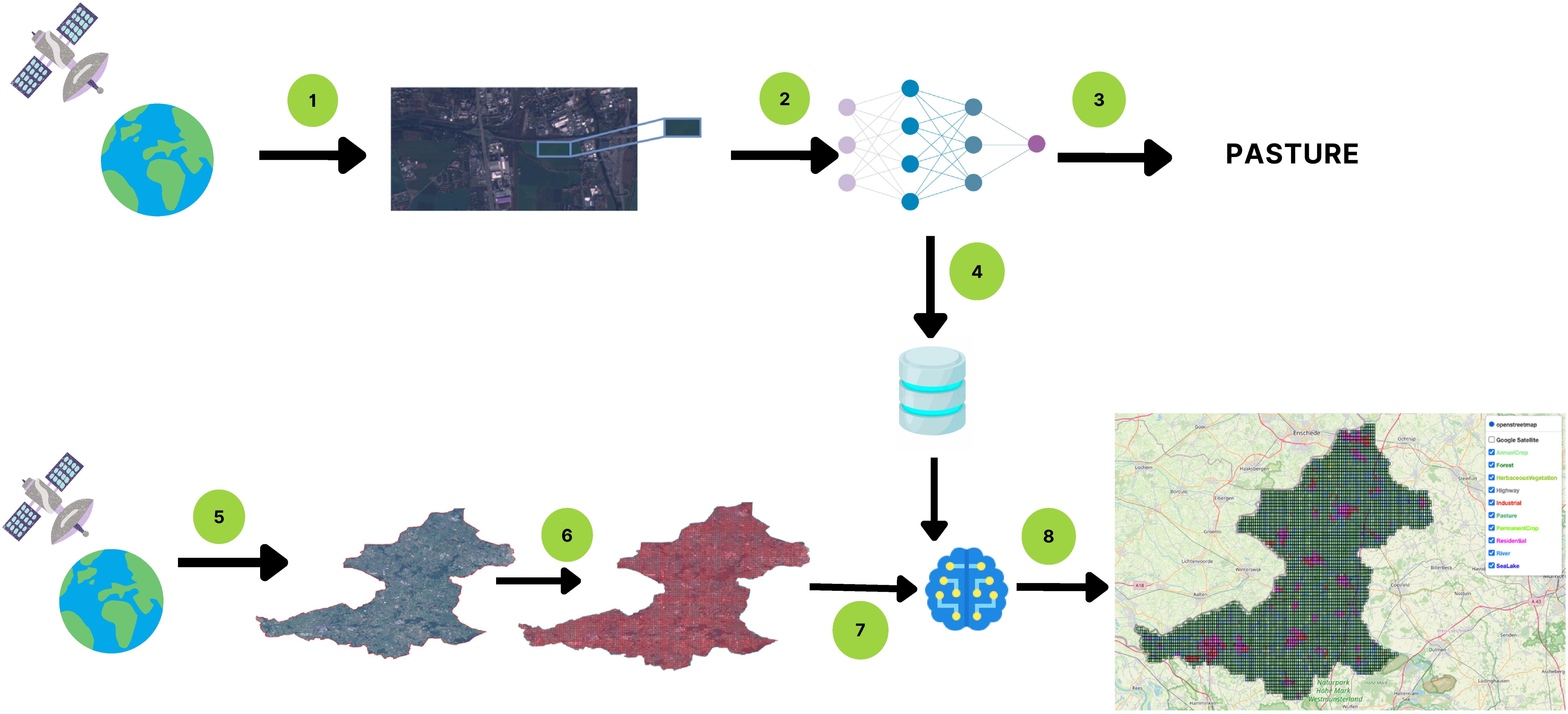}
  \caption{Mapping of LULC using satellite images }
  \label{fig:LULC_Mapping_Process}
\end{figure}

Figure \ref{fig:LULC_Mapping_Process} provides an overview of the LULC classification process using satellite images. Satellites capture images of the Earth. These images are then utilized to extract patches for classification. The objective is to automatically label the physical type of land or its utilization. Each image patch is fed into a classifier, which then outputs the corresponding class depicted on the patch.

However, DL models tend to overfit and demand large quantities of labeled input data to perform well on unseen data \cite{taye_understanding_2023}. This limitation has restricted their adoption in geosciences and remote sensing. Leveraging pre-trained models from optical datasets, such as ImageNet \cite{deng_imagenet:_2009}, can facilitate training new RS models using smaller labeled datasets. Various approaches have leveraged pre-trained models alongside EuroSAT and have demonstrated encouraging results \cite{helber_eurosat:_2019, temenos_interpretable_2023, patel_multi-level_2022}.

With the advancement of Vision Transformers (ViT), many applications are adopting it for image classification tasks \cite{huo_vision_2023}, including EuroSAT \cite{kumari_recent_2023, jannat_improving_2022}. It's suggested that further scaling can enhance performance \cite{zhai2022scaling}, but this model has yet to be integrated with Geospatial data.

The contributions of this paper are summarized below:
\begin{itemize}
\item In this study, we offer a thorough evaluation of the Vision Transformers (ViT) model, considering a range of settings and hyperparameters, with a specific focus on the EuroSAT datasets.
\item We have implemented cutting-edge model improvement techniques to improve the performance of the selected model.
\item The `Kreis Borken' are is mapped using the best-performing model and geospatial data for visual classification.
\end{itemize}
The paper is structured as follows: Section 2 reviews related work. Section 3 describes the dataset used in this study and introduces the methodologies applied using the ViT model. Section 4 presents the results and provides an analysis. Section 5 discusses the findings and concludes the paper.

\section{Related Works}
This section discusses various state-of-the-art image classification techniques that use DL and Transfer Learning (TL) for LULC using the EuroSAT dataset.

\begin{figure}[H]
  \centering
  \includegraphics[width=0.9\textwidth, height=7.0cm]{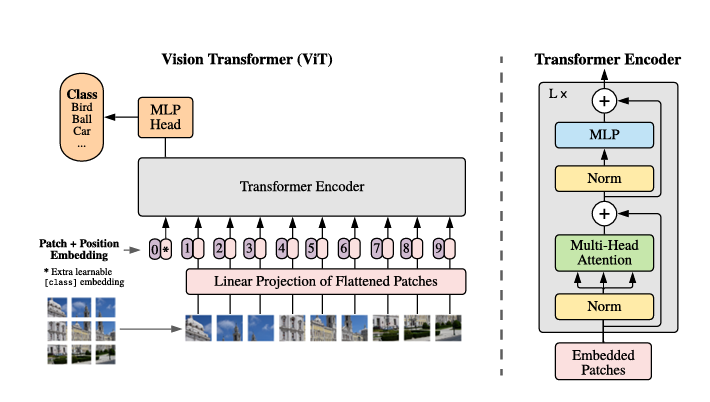}
  \caption{ViT model purposed by Dosovitskiy et al. (2021)}
  \label{fig:ViT_model}
\end{figure}

Fine-tuning large-scale pretrained ViTs has shown prominent performance for computer vision tasks, such as image classification by Dosovitskiy et al. \cite{dosovitskiy_image_2021} using EuroSAT data from its early proposal. The model utilized in his study is shown in Figure \ref{fig:ViT_model}. 

A comparison of various spectral bands—RGB, RGB combined with Near InfraRed (NIR), and multispectral satellite images using the ViT model with Sentinel-2 EuroSAT—was conducted by Anil et al. \cite{Anil2022}. Their experimental results indicate that combining the NIR band with RGB provides more accurate results \cite{anil_influence_2022}. Deep Learning models such as GoogleNet, ResNet-50, Random Forest, VGG, and CNN have been employed for LULC classification.

In their study, Helber et al. \cite{helber2018introducing} tested the GoogleNet and ResNet-50 architectures with various band combinations. Among these, the ResNet-50 using RGB bands outperformed the other configurations. Li et al.'s DDRL-AM method \cite{li_deep_2020} achieved a peak accuracy of 98.74\% with RGB bands. Yassine et al. \cite{yassine_improving_2021} implemented two approaches on the EuroSAT dataset. The first utilized 13 Sentinel-2 spectral bands and achieved 98.78\% accuracy. The second combined these 13 spectral bands with calculated indices, resulting in an improved accuracy of 99.58\%.

Naushad et al. \cite{naushad_deep_2021} advanced LULC classification by employing transfer learning with pre-trained VGG16 and Wide Residual Networks (WRNs) on the EuroSAT dataset's RGB version. Techniques such as data augmentation \cite{shorten_survey_2019}, gradient clipping \cite{zhang_balancing_2023}, adaptive learning rates \cite{ede_adaptive_2020}, and early stopping \cite{dossa_empirical_2021} optimized performance and reduced computational time. Their WRN-based approach achieved a remarkable accuracy of 99.17\%, setting a new benchmark in efficiency and accuracy.

Recently, other variants, including hierarchical ViTs with diverse resolutions and spatial embeddings \cite{he_parameter-efficient_2023}, have been proposed. Without a doubt, the advancements in large ViTs underscore the importance of developing efficient model adaptation strategies.
\section {Materials and Methods}
To classify LULC in a specific region using geospatial data, a transfer learning (TL) task was undertaken. The EuroSAT classes were subsequently color-mapped with the assistance of the ViT model pre-trained on ImageNet-21k. Two datasets were utilized – one with data augmentation and another without. For simplification, the RGB version of the EuroSAT dataset was chosen, and the model was trained using the PyTorch framework. Both model training and testing were conducted using the Tesla T4 GPUs available on Google Colab.

\subsection{Dataset}
\begin{figure}[H]
  \centering
  \begin{subfigure}[b]{0.3\linewidth}
    \includegraphics[width=\linewidth]{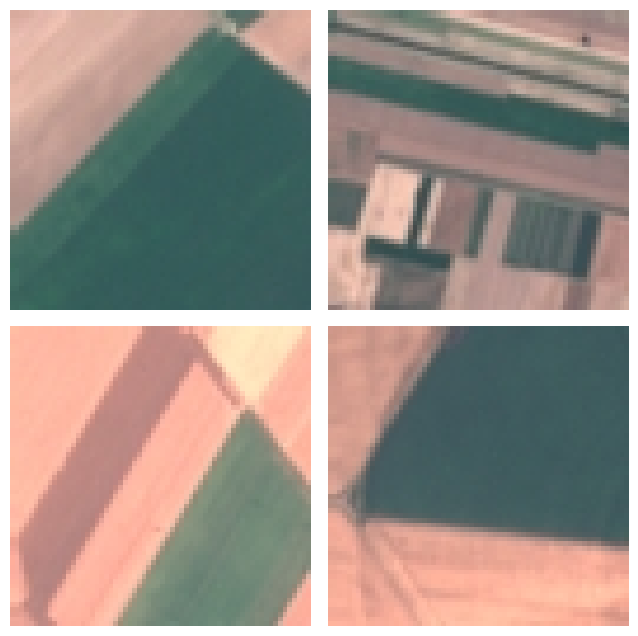}
    \caption{Annual Crop}
    \label{fig:AnnualCrop}
  \end{subfigure}
  \begin{subfigure}[b]{0.3\linewidth}
    \includegraphics[width=\linewidth]{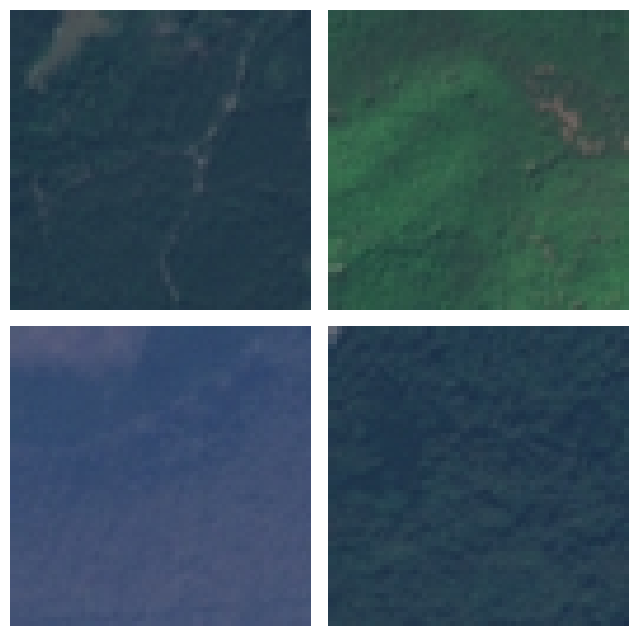}
    \caption{Forest}
    \label{fig:Forest}
  \end{subfigure}
  \begin{subfigure}[b]{0.3\linewidth}
    \includegraphics[width=\linewidth]{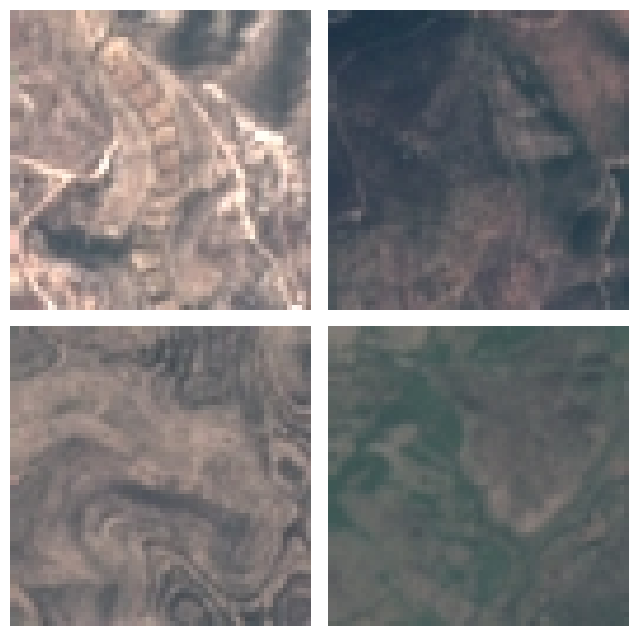}
    \caption{Herbaceous Vegetation}
    \label{fig:HerbaceousVegetation}
  \end{subfigure}
  \caption{Sample example images of three different classes from the EuroSAT dataset.}
  \label{fig:EuroSAT Sample images}
\end{figure}

 The EuroSAT dataset is considered novel and comprises 27,000 labeled and georeferenced images taken from the Sentinel-2 satellite. The images are classified into ten scene classes: Forest, Herbaceous Vegetation, Highway, Pasture, River, Industrial, Permanent Crop, Residential, Annual Crop, and Sea/Lake. Each image patch contains $64\times 64$ pixels with a spatial resolution of 10 m. Figure \ref{fig:EuroSAT Sample images} displays several sample images obtained from the EuroSAT dataset \cite{helber2019eurosat}. The dataset is split into a train set (80\% of data) and a test set (20\% of data) selected at random.

\subsection{Training and Evaluation}
During model training, two sets of data were fed into the system: one with augmentation and one without. In the augmentation process, various image transformation techniques \cite{shijie_research_2017}, such as crops, horizontal flips, and vertical flips, were utilized to augment the data. This strategy aids in preventing the neural network from overfitting to the training dataset, enabling it to generalize more effectively to unseen test data. Since we utilized pre-trained models, the input dataset was normalized to match the statistics (mean and standard deviation) of those models.

The model was trained under various settings, and its accuracy was subsequently evaluated. In this context, the loss was quantified using cross-entropy loss. To counteract potential issues of vanishing or exploding gradients during training, which could adversely affect the parameters, the gradient clipping technique \cite{zhang_why_2020} was employed with a value set to $1.0$. We used the Adam optimizer, combined with multiple learning rates, due to its proven efficacy in image classification tasks \cite{10151232}. Regularization strategies, including early stopping, dropout, and weight decay \cite{8864616}, were also implemented to combat overfitting and to optimize time and resources. We also recorded the total duration of the experiment \footnote{https://github.com/sumn2u/LULC-Mapping.git}.

\section{Results}
This section presents the results obtained from ViT, VGG16, and ResNet-50 models with same settings and narrows down the ViT model performance under different settings. Metrics such as accuracy and the time taken to train the model are measured. The model's performance is assessed using both data augmentation techniques and without data augmentation, maintaining similar settings for both.

\begin{table}[H]
  \centering
  \caption{Comparative experimental results of ViT, ResNet-50 and VGG16 models with and without data augmentation.}
  \begin{tabular}{cccc}
    \toprule
    Model & Epoches Trained & Total Time & Accuracy \\
    \midrule
    ViT (Non Augmented Data) & 15 & 12427.12 sec & 98.61\% \\
    ViT (Augmented Data) & 15 & 11045.71 sec & 98.72\% \\
    VGG16 (Non Augmented Data) & 15 & 5070.30 sec & 98.06\% \\
    VGG16 (Augmented Data) & 15 & 4996.54 sec & 97.78\% \\
    ResNet-50 (Non Augmented Data) & 15 & 3602.74 sec & 98.52\% \\
    ResNet-50  (Augmented Data) & 15 & 3329.52 sec & 98.48\% \\
    \bottomrule
  \end{tabular}
  \vspace{2pt} 
 \label{tab:Comparative_model_experimental_results}
\end{table}

Table \ref{tab:Comparative_model_experimental_results} shows that the ViT model is more accurate in Augmented and Non-Augmented Data, but takes longer to train than other models. Conversely, the ResNet-50 model exhibits better accuracy than VGG16, and it requires relatively less training time for both augmented and non-augmented data.

\begin{table}[H]
  \centering
  \caption{Comparative experimental results of ViT model with and without data augmentation.}
  \begin{tabular}{cccc}
    \toprule
    Model & Epoches Trained & Total Time & Accuracy \\
    \midrule
    ViT (Non Augmented Data) & 10 & 11045.71 sec & 98.67\% \\
    ViT (Non Augmented Data) & 15 & 12427.12 sec & 98.61\% \\
    ViT (Non Augmented Data) & 20 & 15604.66 sec & 98.41\% \\
    ViT (Augmented Data) & 10 & 7356.24 sec & 98.24\% \\
    ViT (Augmented Data) & 15 & 11045.71 sec & 98.72\% \\
    ViT (Augmented Data) & 20 & 14377.87 sec & 99.19\% \\
    \bottomrule
  \end{tabular}
  \vspace{10pt} 
 \label{tab:Comparative_experimental_results}
\end{table}

\begin{figure}[H]
  \centering
  \includegraphics[width=0.98\textwidth, height=5.0cm]{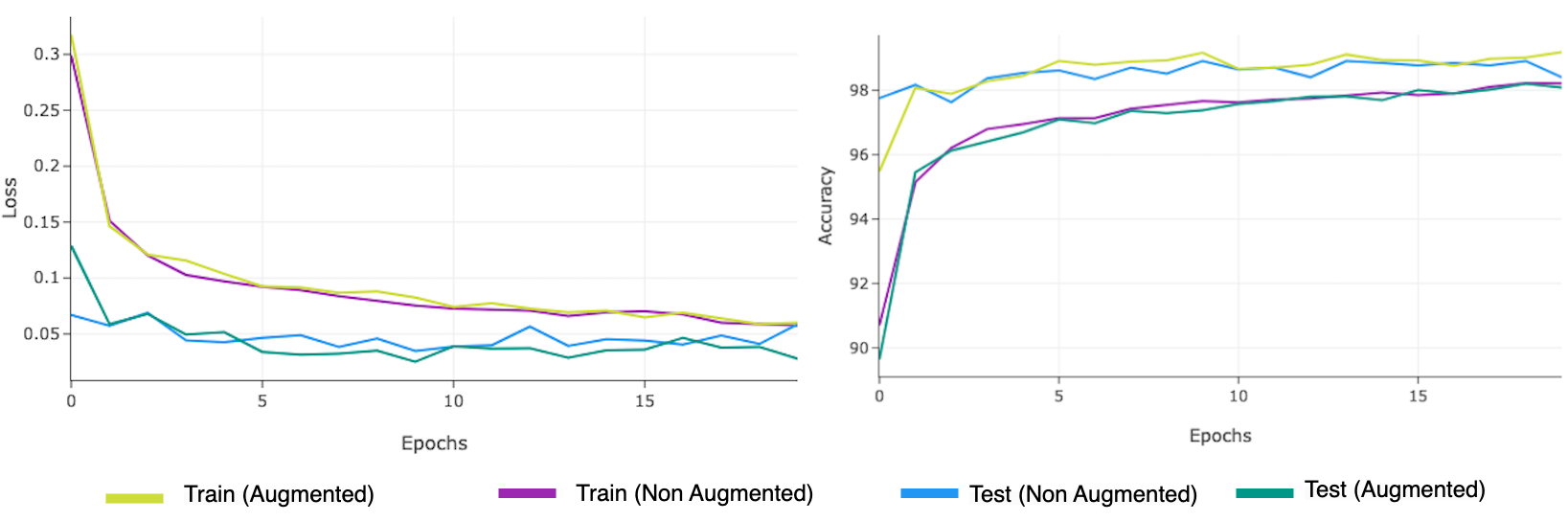}
  \caption{Comparison of loss and accuracy of models at different epochs and iterations.}
  \label{fig:loss_accuracy_comparison}
\end{figure}

The accuracy of the model increases as it trains longer using augmented data; in contrast, the accuracy decreases for the non-augmented version, as observed in Table \ref{tab:Comparative_experimental_results}. The training time for the augmented data is also relatively longer compared to the non-augmented data. The loss and accuracy of both the augmented and non-augmented models are depicted in Figure \ref{fig:loss_accuracy_comparison}. 
\begin{figure}[H]
  \centering
  \begin{subfigure}[b]{0.42\linewidth}
    \includegraphics[width=\linewidth]{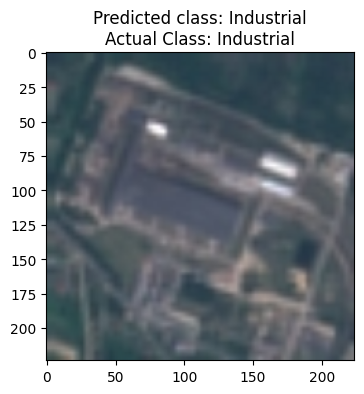}
    \caption{Data Augmentation (20 epochs)}
    \label{fig:prediction_20_industrial}
  \end{subfigure}
  \begin{subfigure}[b]{0.42\linewidth}
    \includegraphics[width=\linewidth]{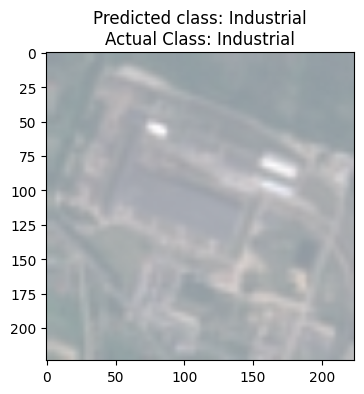}
    \caption{Without Data Augmentation (20 epochs)}
    \label{fig:prediction_20_industrial_non_aug}
  \end{subfigure}
  \caption{ Both augmented with 20 epochs and non augmented with 20 epochs model was able to predict correct results with test data }
  \label{fig:Exprimental Results}
\end{figure}
\begin{figure}[H]
  \centering
  \begin{subfigure}[b]{0.49\linewidth}
    \includegraphics[width=\linewidth]{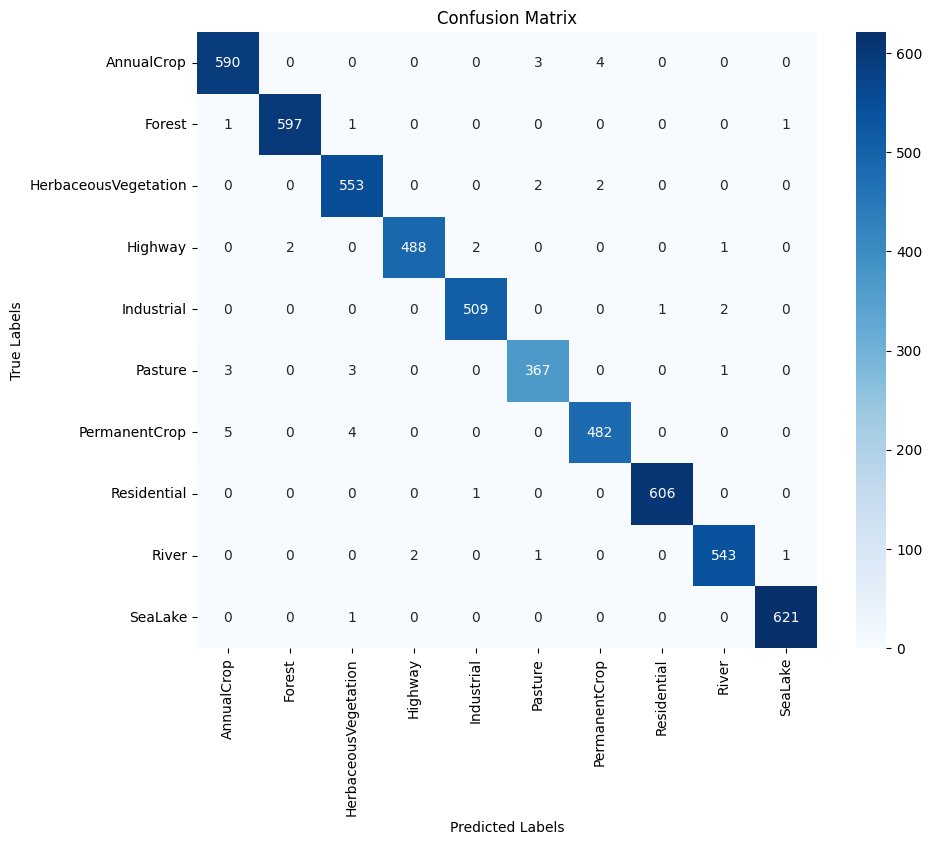}
    \caption{Confusion Matrix (Augmented)}
    \label{fig:confusion_matrix_aug_20_epoch}
  \end{subfigure}
  \begin{subfigure}[b]{0.49\linewidth}
    \includegraphics[width=\linewidth]{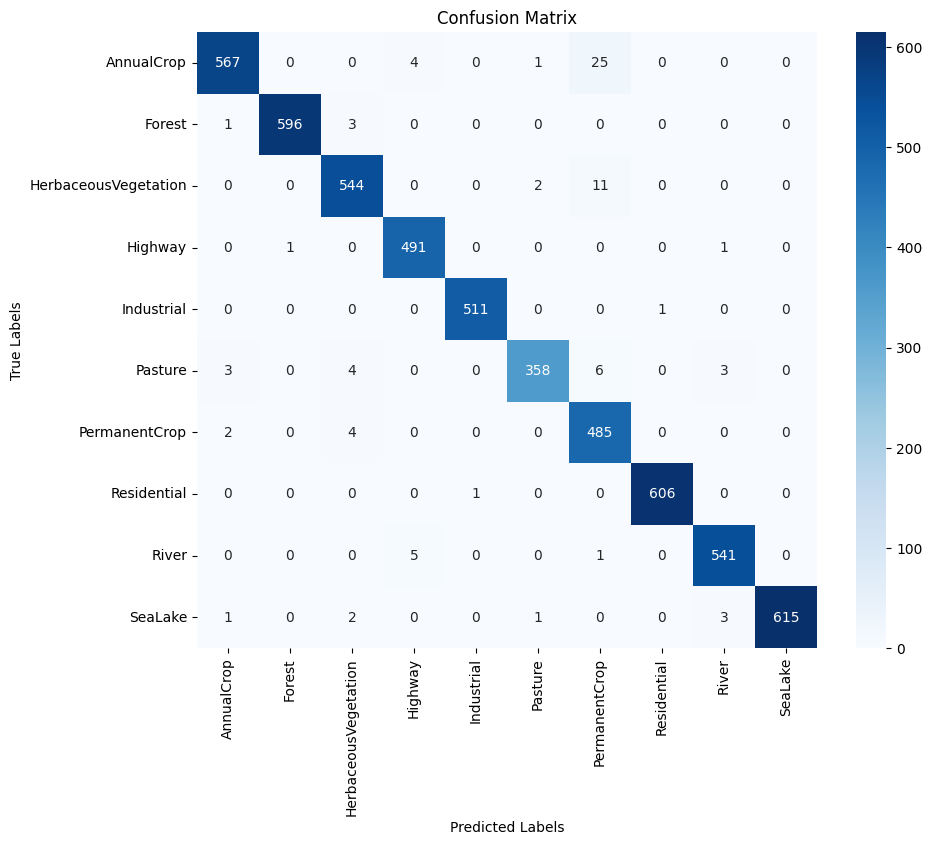}
    \caption{Confusion Matrix (Non Augmented)}
    \label{fig:confusion_matrix_no_aug_20_epoch}
  \end{subfigure}
  \caption{Confusion matrix of augmented and non augmented data with ViT model for 20 epochs}
  \label{fig:Confusion Matrix}
\end{figure}
Subsequently, the model is tested with test data, as shown in Figure \ref{fig:Exprimental Results}. Figure \ref{fig:Confusion Matrix} shows the confusion matrix of ViT model with and without data augmentation on validation data. The Forest and Sea Lake classes showed best performance with higher accuracy in augmented data.  The Pasture class has least accuracy mean while the rest has almost $99$\% accuracy.
\begin{figure}[H]
  \centering
  \includegraphics[width=0.97\textwidth, height=8cm]{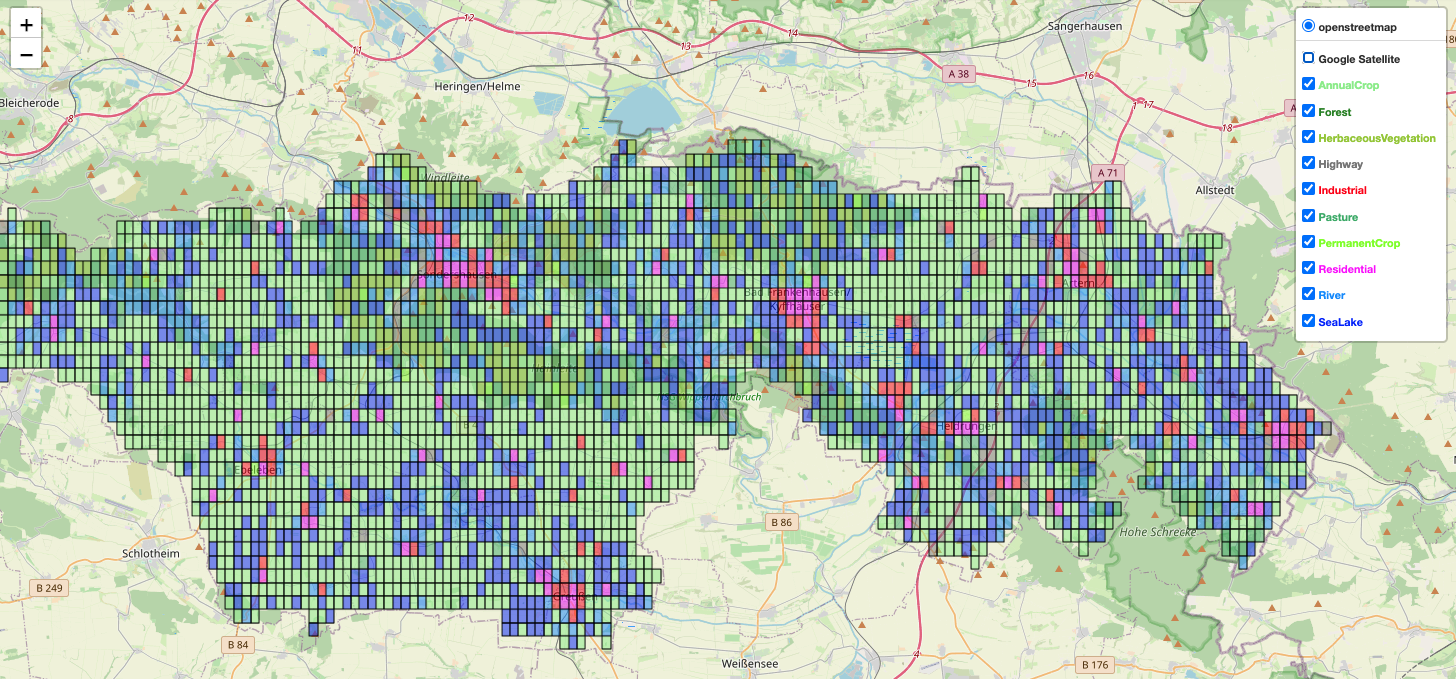}
  \caption{Mapping of LULC of `Kreis Borken' area using satellite images and ViT model.  }
  \label{fig:LULC_Mapping_Kreis_Borken}
\end{figure}
From the experiments conducted, we found that the model trained with augmented data yields better accuracy. Using Google Earth Engine and Sentinel 2A \cite{9665519}, we selected the Kreis Borken area. The data, drawn from satellite images spanning 2018 to 2020, was segmented into $64\times 64$ tiles within the specified boundary. Using the model, these tiles were classified and color-mapped, as depicted in Figure \ref{fig:LULC_Mapping_Kreis_Borken}.

\section{Conclusion}
The objective of this research was to explore the application of transfer learning in LULC classification. We utilized the ViT model, pre-trained on ImageNet-21K, and fine-tuned it with the RGB bands of the EuroSAT dataset for classification. Consistent with other experiments, the results of this study suggest that transfer learning is a reliable method capable of delivering superior results. Our approach advanced the state-of-the-art by achieving a $99.19$\% accuracy for the RGB bands of the EuroSAT dataset. We also compared the classification results with and without data augmentation. Data augmentation was observed to enhance the visual variability of each training image without introducing new spectral or topological information, thereby enriching the dataset's diversity. The performance of the augmented data surpassed that of the model trained on the original dataset.

Additionally, this study incorporated model enhancement techniques, including regularization, early stopping, gradient clipping, and learning rate optimization, to optimize model training, improve performance, and reduce the computational time needed. The most proficient model was subsequently used to map class distributions and offer insights into a specific region of geospatial imagery. Such insights can assist in monitoring shifts in land usage and shaping policies for environmental conservation and urban development. 
\bibliographystyle{unsrt}
\bibliography{tackling_climate_workshop}

\end{document}